# DEEVA: A Deep Learning and IoT Based Computer Vision System to Address Safety and Security of Production Sites in Energy Industry

**Nimish M. Awalgaonkar**
School of Mechanical Engineering, Purdue University
nawalgao@purdue.edu

**Haining Zheng**
ExxonMobil Research and Engineering Company
haining.zheng@exxonmobil.com

**Christopher S. Gurciullo**
ExxonMobil Research and Engineering Company
chris.s.gurciullo@exxonmobil.com

**Abstract**

When it comes to addressing the safety/security related needs at different production/construction sites, accurate detection of the presence of workers, vehicles, equipment, materials etc. is of prime importance and has therefore formed an integral part of computer vision-based surveillance systems (CVSS). In past, traditional CVSS systems focused on the use of different computer vision and pattern recognition algorithms which were overly reliant on manual extraction of features and small datasets, limiting their usage because of low accuracy, need for expert knowledge and high computational costs. In order to address these limitations, the main objective of this paper is to provide decision makers at sites with a practical yet comprehensive deep learning and IoT based solution to tackle various computer vision related problems such as scene classification, object detection in scenes, semantic segmentation, scene captioning etc. Our overarching goal is to address the central question of "What is happening at this site and where is it happening?" in an automated fashion – minimizing the need for human resources dedicated to surveillance. We developed Deep ExxonMobil Eye for Video Analysis (DEEVA) package to handle scene classification, object detection in scenes, semantic segmentation and captioning of scenes in a hierarchical approach. The results reveal that transfer learning with the RetinaNet object detector is able to detect the presence of workers, different types of vehicles/construction equipment, safety related objects at a high level of accuracy (above 90%). With the help of deep learning to automatically extract features and IoT technology to automatic capture, transfer and process vast amount of realtime images, this framework is an important step towards the development of intelligent surveillance systems aimed at ad-dressing myriads of open ended problems in the realm of security/safety monitoring, productivity assessments and future decision making.

## 1 Introduction

Over the years, several computer-vision related tools have been developed by researchers and engineers to automatically detect objects of interest (e.g. detection of presence of workers, different types of vehicles/construction equipment, safety related objects etc.) from images or videos on concerned production/construction sites in order to improve the safety, security and productivity across these sites. In such a context, specific applications of computer-vision include: detection and tracking of workers, vehicles, and the use of safety equipment etc.

The main objective of this paper is to provide users with a practical yet comprehensive deep learning based package to tackle various computer vision related problems such as scene classification, object detection in scenes, semantic segmentation, scene captioning with IoT devices at remote oil & gas production site.

This paper is organized as follows. In Section 2, we discuss related work. In Section 3, we formulate the problem and present our methodologies. In Section 4, we show results we have obtained and performance metrics. Section 5 summarizes our findings and concludes the paper.

## 2 Related Work

Traditional computer-vision algorithms used to exploit output frames from video cameras in conjunction with feature engineering methods such as the histograms of oriented gradients (HoG), histogram of optimal flow (HoF), scale invariant feature transform (SIFT) etc. to infer certain quantities of interest/semantic information about the concerned sites of interest. However, these methods tend to be overly reliant upon manually extracting hand-crafted features from input frames using conventional machine learning/pattern recognition algorithms. Constructing feature extractors for such applications require considerable domain expertise and careful engineering to extract meaningful feature vectors from the input frames data which are then used as inputs by a classifier (e.g., Logistic Regression, Support Vector Machines etc.) to infer concerned quantities of interest (e.g. label associated with an entire image in the context of scene classification problem, labels and bounding boxes in the context of object detection etc.) Therefore, with such methods, the performance of the data-driven model is highly dependent upon designing an effective feature extractor that is able to appropriately characterize high level semantic features associated with the concerned images/frames. Moreover, due to high computational cost associated with feature extraction, these models were typically trained using small datasets (<5k images/label), which limits the intra and inter class variability the model is trained over. This further hinders the ability of these models to be generalizable and thereby be able to accurately infer certain quantities of interest associated with video frames of concerned sites.

To address this limitation of extracting hand-crafted features from high-dimensional input images, we make use of Deep Convolutional Neural Networks (DCNNs), an alternative end-to-end solution to automatically infer concerned quantities of interest without the need for domain expertise based feature extraction. DCNNs consist of multiple convolutional and pooling layers interspersed with non-linearity layers to extract high-level semantic features characterizing the input image/frame. While several studies in the field of construction engineering have explored the use of DCNNs for addressing problems such as detecting safe/unsafe behavior of workers, pose detection, detection of construction equipment etc., they were based on architectures that were developed 4-5 years ago and are not state-of-the-art anymore. For example, in the context of safety related studies, Fang et al. made use of Faster-R CNN object detector to detect workers and their concerned personal safety equipment (Fang et al. 2018). In the context of detection and tracking of workers, Ding et al. made use of a hybrid deep learning based model to detect dangerous events happening in the concerned video frames (Ding et al 2019). In the context of large construction equipment detection, Kim et al. employed R-FCN object detector (fine-tuned on their own labeled dataset using transfer learning) to detect a variety of construction equipment (Kim et al. 2018). All of these papers tackle the important problem of object detection. By being able to detect workers, safety equipment, construction equipment etc, unsafe behavior and associated site conditions can be automatically detected, thereby providing owners and decision makers across different sites with a mechanism to improve safety/security issues at their site.

# 3 Methodology

Our overarching goal is to help decision makers across different production/construction sites be able to address the central real-time computer vision and NLP processing question of "What is happening at this site and where is it happening?". We build up towards this goal in a hierarchical fashion, from the label density point of view from scene classification to object detection in scenes and to semantic segmentation and to captioning of scenes on the other axis of label complexity. Combing these two dimensions, more complex tasks like Region based cognitive captioning can be achieved, e.g. Excavator is standing still on the ground, two engineers wearing hard-hats are standing in front of an excavator, engineers have some kind of tool in their hands and engineers are wearing required protection gears.

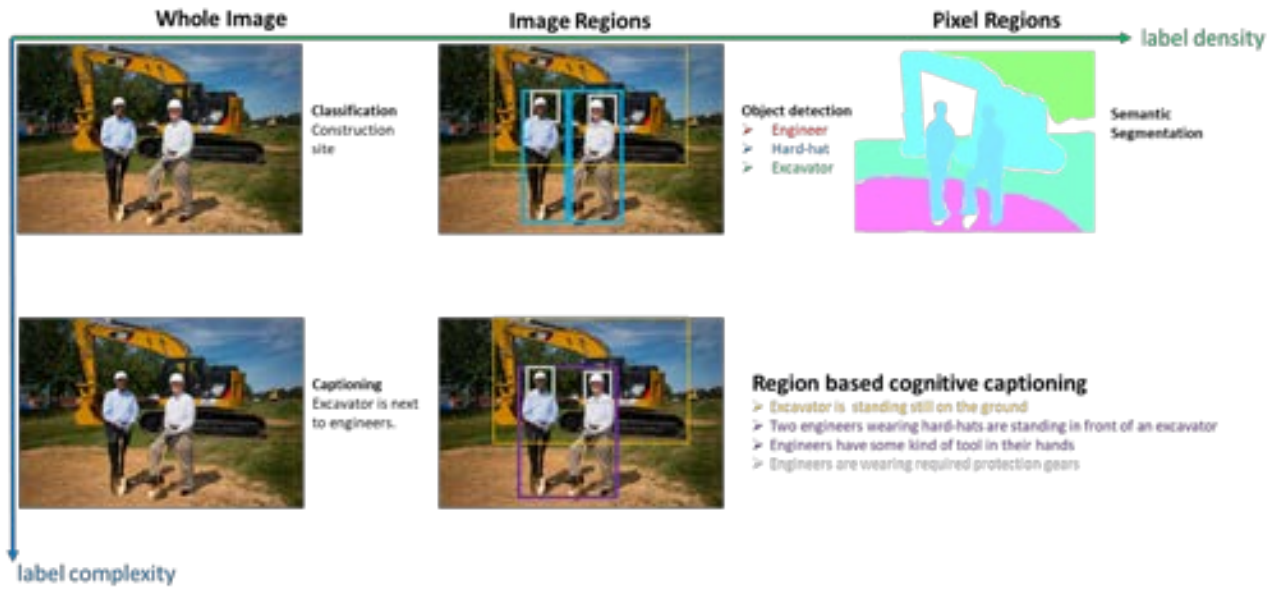

Fig. 1. Hierarchical Approach with Constructing Necessary Lego Blocks. Picture from Business Wire.

First, engineers need to decide what task they want to address, e.g. classification, object detection? Then they can determine the necessary architectures, layers and loss functions as Lego blocks (Table 1). If Lego block is black in color, DEEVA provides engineers with the following color options/ architectures in Table 2.

Table 1. Necessary Lego blocks for Various Tasks

| Tasks | |
|---|---|
| Whole Image Classification | |
| Object detection | |
| Semantic Segmentation | |
| Captioning | |

Table 2. Architecture/Module Options for Black Lego Block

| Modules | | | | |
|---|---|---|---|---|
| Backbone architectures | ResNet | SqueezeNet | DenseNet | |
| Detector layers | RetinaNet | YoloV3 | TinyYoloV3 | |
| Decoder architectures | RetinaNet | YoloV3 | TinyYoloV3 | |
| Loss Functions | Cross-entropy loss | Focal loss | L1 loss | L2 loss |

## 3.1 Data Preparation and Labeling

This section is devoted to the first step in any data-driven modeling scheme: collecting and labeling relevant datasets to address the open-ended questions at hand. There is a paucity of publicly available images of objects from construction/production sites (e.g., workers, plant, equipment, and materials), which has inhibited the development of effective intelligent monitoring systems. To demonstrate the robustness of the method proposed in this paper, it is necessary to establish a dataset of images of objects to enable the detection of different kinds of activities on-site. In order to address this problem and also avoid potential bias, we made use of a wide-array of publically available datasets in order to train our models. The images in the dataset were collected from different viewpoints, at varying scales, poses, occlusions and under changing lighting conditions.

- AIM Construction dataset (Kim et al. 2018), 3200 images
- MS-COCO dataset(Lin et al. 2014), 120,000 images
- KITTI-TCO Surveillance dataset (Geiger et al. 2012), 150,000 images
- Hard-hat dataset (Wu et al. 2019), 3174 images
- Google Open Images dataset (Kuznetsova et al. 2018), 6000 images
- CityScapes video dataset (Cordts et al. 2016), 25,000 frames

## 3.2 Exploiting Different Deep Architectures

### 3.2.1 Backbone Structure: ResNet

The power of many convolution neural networks have been shown through utilizing the ImageNet dataset. When an image is fed into such a network, the results of its last a few layers would be an effective representation of the input. Thus we extract feature maps with a backbone whose parameters are initialized with the pre trained model on ImageNet dataset.

### 3.2.2 Object Detector: RetinaNet

A RetinaNet object detector is built on top of ResNet backbone (He et al. 2016) structure by making two improvements over existing single stage object detection models, like YOLO (Redmon et al. 2016) and SSD (Liu et al. 2016): feature pyramid networks (FPN) for object detection and focal loss for dense object detection.

3.2.2.1 Feature Pyramid Network

We adopt the Feature Pyramid Network (Lin et al. 2017-1) to extract further pyramid features. The structure we use is similar as the structure described in RetinaNet. (Lin et al. 2017-2). The structure can be seen in Fig. 2, where (b) are the outputs of 3 layers of the backbone ResNet (a). With the Feature Pyramid Network (FPN), we can get pyramid features. Higher level feature maps contain grid cells that cover larger regions of the image and is therefore more suitable for detecting larger objects; on the contrary, grid cells from lower level feature maps are better at detecting smaller objects.

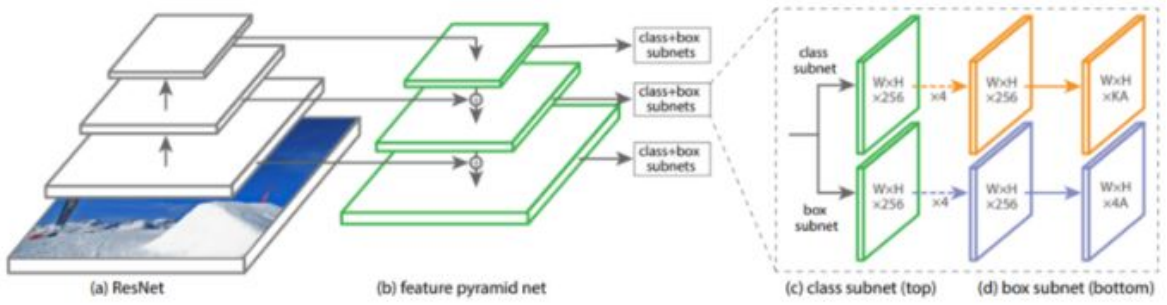

Fig. 2. Object Detector Architecture

With the help of the top-down pathway and lateral connections, which do not require much extra computation, every level of the resulting feature maps can be both semantically and spatially strong. These feature maps can be used independently to make predictions and thus contributes to a model that is scale-invariant and can provide better performance both in terms of speed and accuracy. To this backbone, two subnetworks are attached: one for classifying anchor boxes (c) and one for regressing from anchor boxes to ground truth boxes (d). The proposal boxes are generated based on the shapes of these features, as described in the next subsection.

3.2.2.2 Anchors

Anchors are the proposed boxes for objects. As mentioned above, the anchors are generated according to the shape of pyramid features. For each of the pixels, we generate an anchor whose height and width are both 2x, and then we shift their centers into the corresponding pixels of the input. Furthermore, we expand each anchor into 9 with three aspect ratios and three anchor resize scales. However, with these expansions, there may be some anchors whose centers are located out of the input, so a filter is deployed to delete these illegal anchors. These operations are operated on all of the pyramid features. As a result, we have many dense anchors, which should be able to cover the objects in videos.

3.2.2.3 Regression Subnet

The regression subnet is aimed at regressing the offset of anchors to nearby ground-truth objects. This subnet is a combination of 2D Convolutional and Convolutional LSTM layers. It takes advantage of deep CNNs without making the model too complex. Except the last layer, the number of filters in each layer is 64. The number of filters in the last layer is 9×4 for each of the four boundaries of the 9 anchors of a given pixel. The offset of the boundaries is set as the settings in R-CNN (Girshick 2015). The regression subnet is connected to the regression loss function with L1 regularization in the training model.

3.2.2.4 Classification Subnet

The classification subnet is deployed to predict the probability of object presence for each of the anchors and object classes. The structure of classification subnet is the same as regression subnet, as can be seen in Fig. 2. The number of filters in each layer except the last one is also 64. The last layer, whose number of filters is set as 9× the number of classes, is responsible for predicting the probability for each of the 9 anchors centered at a given pixel in the feature map containing each of the classes, so it is activated by a sigmoid function. The classification subnet is further connected to the focal loss function in the training model which is as defined below.

3.2.2.5 Focal Loss

Without much preprocessing, one-stage detection methods usually generate proposal boxes regardless of the values in the three channels of images, so most of these proposals may locate in the background area. If we regard the loss for background and objects equally, the model will tend to classify each box as background. Lin et al. (Lin et al. 2017-2) proposed a focal loss for objection detection on images to solve this problem. The α-balanced form of focal loss is defined as:

$$FL(p,y) = \begin{cases} -\alpha(1-p)^{\gamma}\log p, & y=1 \\ -(1-\alpha)p^{\gamma}\log(1-p), & y=0 \end{cases} \quad (1)$$

where $y \in \{0, 1\}$ specifies the ground-truth class in one-hot and $p \in [0, 1]$ is the probability that model estimated for the class with label $y = 1$. This loss function reduces the loss compared to cross entropy when $\gamma > 0$, but it makes the loss for well-classified samples relatively smaller than those $p < 0.5$, preventing the model to misclassify boxes with objects as background.

## 3.3 Transfer Learning

In practice, very few researchers train an entire Convolutional Network (ConvNet) from scratch (with random initialization), because it is relatively rare to have a dataset of sufficient size. Instead, it is common practice to pre-train a ConvNet on a very large dataset (e.g. ImageNet), and then use the ConvNet as a fixed feature extractor for the task of interest. Depending on the size of available dataset, we either re-initialize the last single layer (small dataset) or the last a few layers (large dataset), then freeze the parameters of other layers and only train the parameters of last single or a few layers, respectively. AlexNet is used for illustration purposes in Fig. 3 for this process but same applies for other backbones.

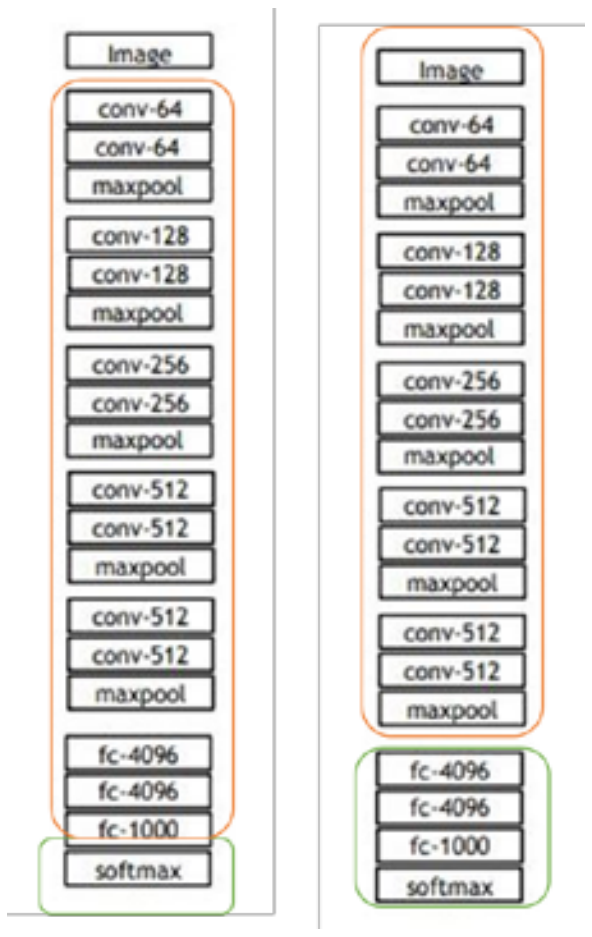

Fig. 3. Transfer Learning with Single Layer Freezing for Small Dataset (Left) and Multiple Layer Freezing for Large Dataset (Right)

## 3.4 Semantic Segmentation

The unmanned production site is interested to know if the gate is open or closed and if there are any discontinuities in the fencing structure, possibly due to damage caused by intruders. Different from object detection task, rather than bounding boxes, we are now interested in diving deepsemantic understanding at finer pixel level and classify each pixel in the image from a pre-defined set of classes. We follow an encoder/decoder structure for semantic segmentation modeling, where we downsample the spatial resolution of the input, developing lower-resolution feature mappings which are learned to be highly efficient at discriminating between classes, and the upsample the feature representations into a full-resolution segmentation map.

## 3.5 Scene Captioning

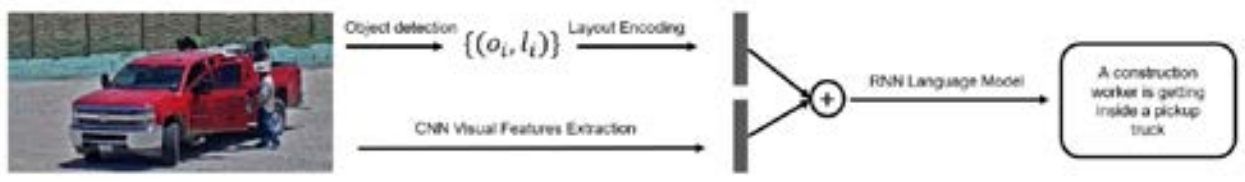

Fig. 4. Scene Captioning Architecture

We exploit the framework in Fig. 4 for the scene captioning task, which captures the structural semantic within visual scene by RNN language model with inputs from object detection/layout encoding and visual feature extraction.

## 3.6 Hyper-parameter Optimization

Deep Learning is extremely powerful, yet tuning Deep Learning models is enormously non-intuitive. If the whole deep neural net is noted as a function g, the challenges are that $g$ is explicitly unknown, expensive, non-convex and no gradient information is available. A practical approach to optimize deep neural net hyper-parameter using Bayesian modelling can be found in Fig. 5.

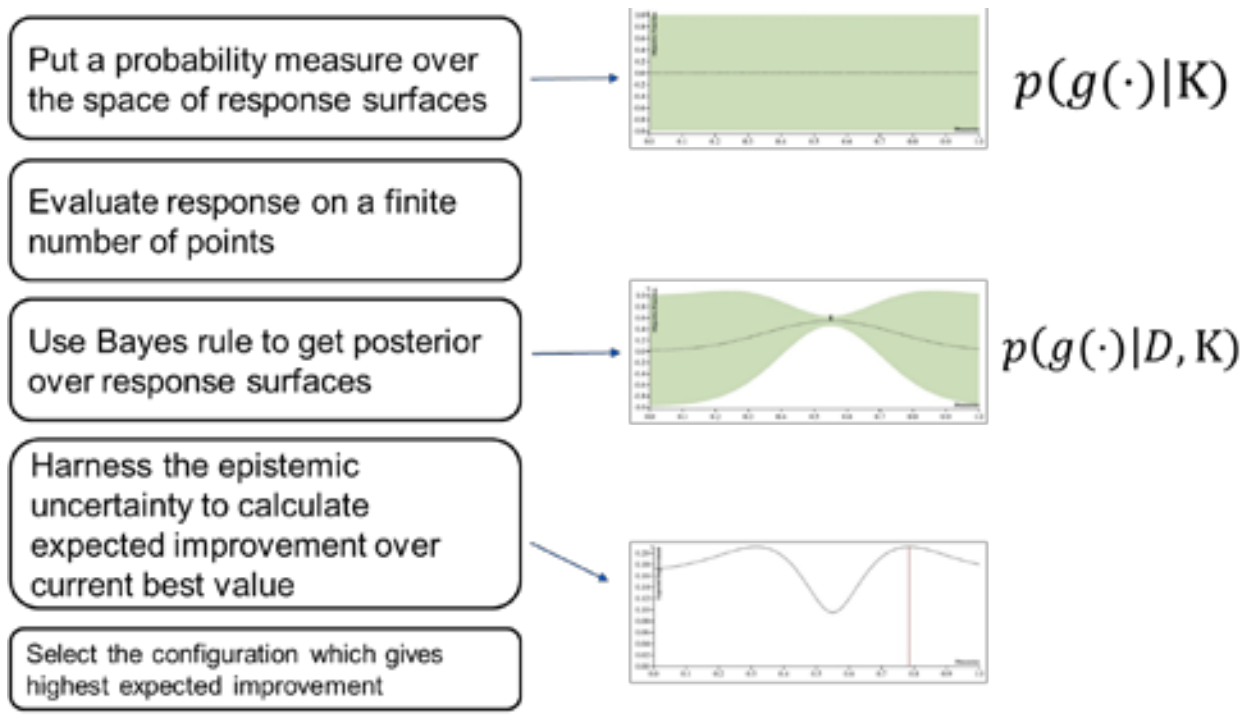

Fig. 5. Optimization under Uncertainty using Gaussian Processes

## 4 Results Analysis

### 4.1 Construction Site Scene Objective Detection

With the DEEVA objective detection framework and training data set discussed in 3.1.1, we were able to use transfer learning to retrain the last layers of RetinaNet and achieved very decent results, as shown in in Fig. 6. Not only our package is able to detect the objects in the image with high accuracy, including equipment like excavators, excavator buckets, and people, but we also are able identify the safety hats, safety glasses and tools on the engineer's hands. The model training, validation loss and mean average precision during model training is given in Fig. 7.

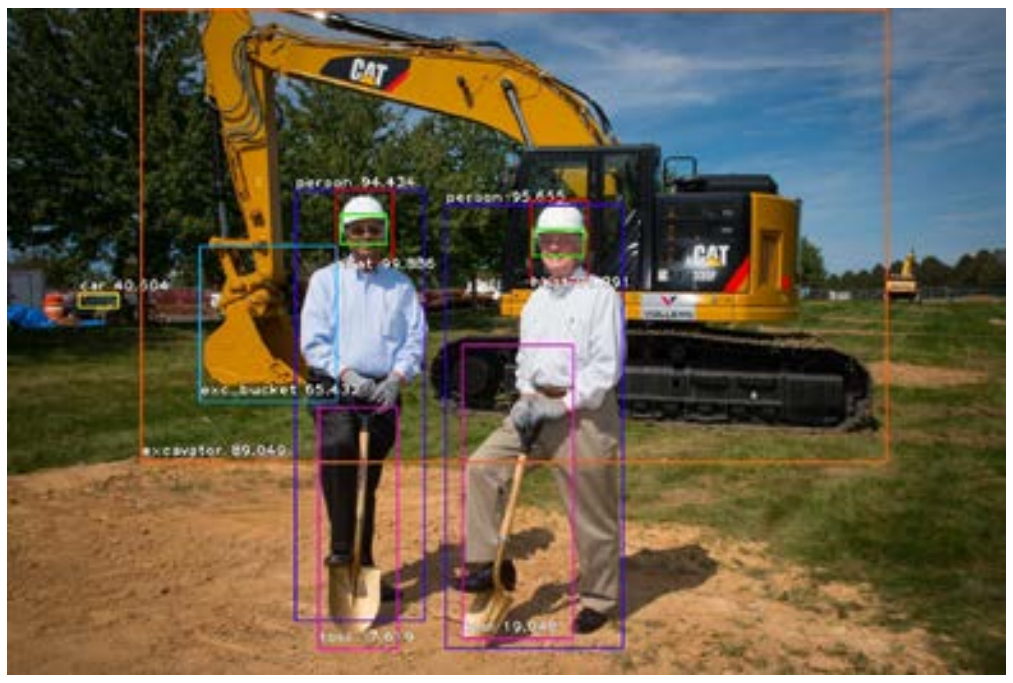

Fig. 6. Example Objective Detection Results on a Construction Site

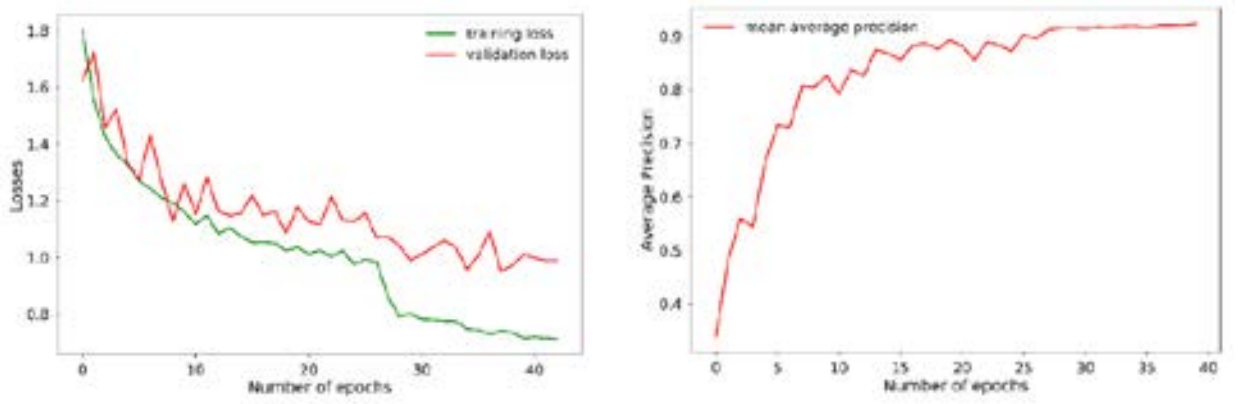

Fig. 7. Training, Validation Loss and Mean Average Precision during Model Training

### 4.2 Construction Equipment Objective Detection

We also tested the DEEVA package on the construction equipment dataset use in (Kim et al. 2018). We divided the training, validation and testing set into 70:15:15 and achieved pretty good precision after 10 to 20 epochs as shown in Fig. 8. The test precision for the 6 equipment (mostly above 90%) can be found in Table 3.

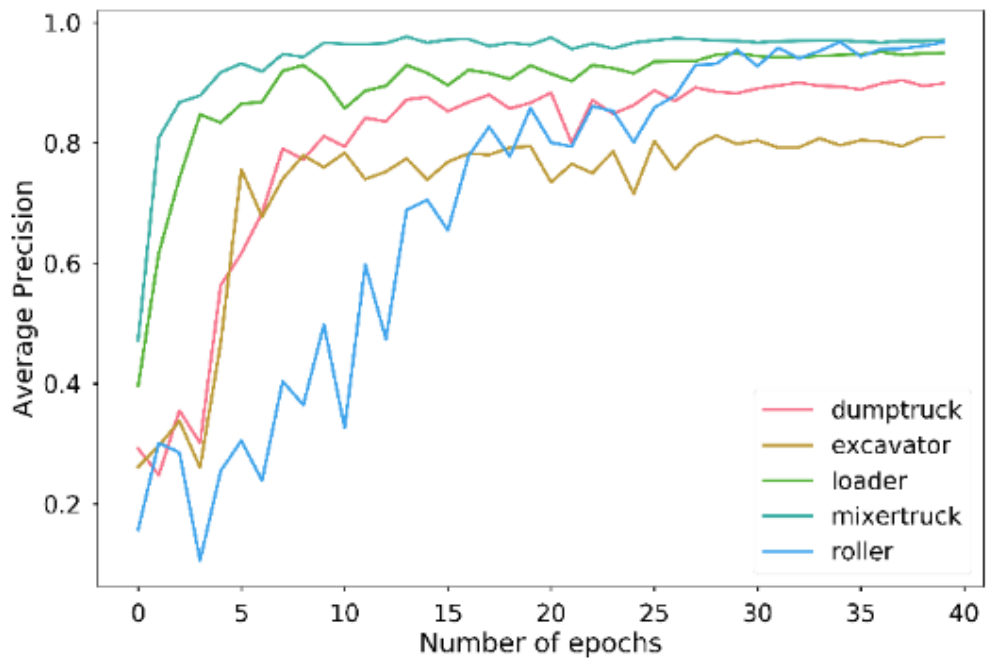

Fig. 8. Precision for Various Construction Equipment during Training

Table 3. The Test Precision for Various Equipment

| Classes | Dump-truck (123) | Excava-tor (62) | Loader (121) | Mixer-truck (103) | Roller (54) | mAP |
|---|---|---|---|---|---|---|
| Perfor-mance metrics | 89.92% | 80.86% | 95.06% | 97.12% | 97.04% | 92.48% |

### 4.3 Hyper-parameter Optimization Results

As discussed in section 3.6, we developed a Bayesian Hyper-parameter optimization model using Gaussian Processes. We were able to improve the mean average precision (mAP) from 0.65 to 0.74 and saved at least 20 hours of computation time. As illustrated by the green curve in Fig. 9, we achieved mAP of 0.74 at the 3rd iteration while the random search approach only achieved mAP of 0.65 after 8 iterations.

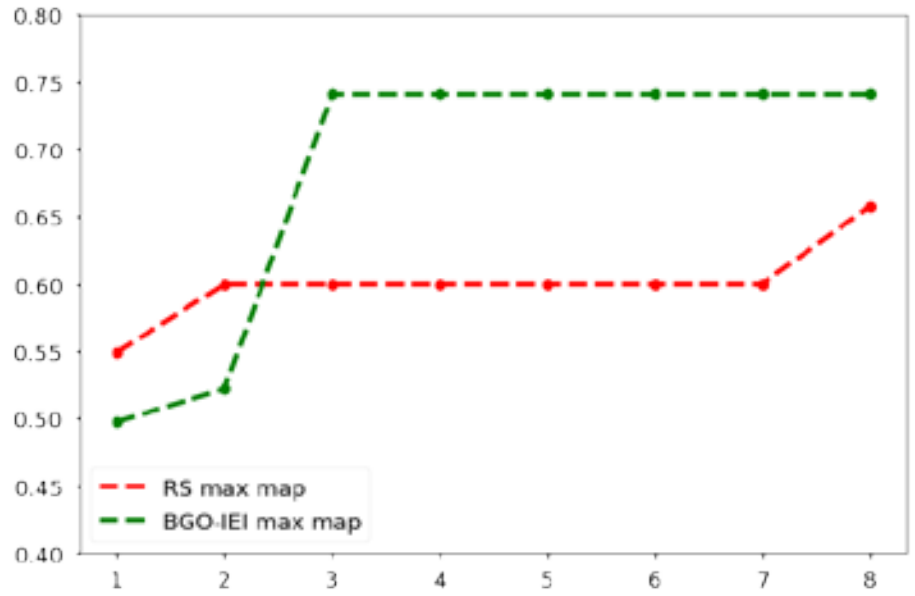

Fig. 9. Sample Results of Hyper-parameter Optimization

## 5. Conclusions

In this paper, we introduced DEEVA (Deep ExxonMobil Eye for Video Analysis), a deep learning and IOT based computer vision system to process computer vision and natural language in real time in order to address the safety and security of production sites in energy industry. We developed the system in a hierarchical fashion in two separate directions of label complexity and label density. Our results reveal that transfer learning with DEEVA is able to detect the presence of workers, different types of vehicles/construction equipment, safety related objects at a high level of precision (above 90%). This work builds the foundation for a fully intelligent surveillance systems with AI as its brain to make decisions and IoT as the digital nervous system to acquire video, image, sound and thermal data.

## Acknowledgments

The authors would like acknowledge helpful discussions with Jeffery Ludwig and Antonio Paiva and support from Robert A. Johnson.